\documentclass[sigconf]{acmart}
\copyrightyear{2027}
\acmYear{2027}
\setcopyright{cc}
\setcctype{by}
\acmConference[ASPDAC '27]{32nd Asia and South Pacific Design Automation Conference}{January 25--28, 2027}{Tokyo, Japan}
\acmBooktitle{32nd Asia and South Pacific Design Automation Conference (ASPDAC '27), January 25--28, 2027, Tokyo, Japan}
\acmDOI{10.1145/3840404.3848291}
\acmISBN{979-8-4007-2923-2/2027/01}

\AtBeginDocument{%
  }

\usepackage{tabularx}
\usepackage{enumitem}
\usepackage{xcolor}

\newcommand{\tblue}[1]{\textcolor{black}{#1}}

\begin{document}

\author{
Quanling Zhao\textsuperscript{1},
Ava Emami\textsuperscript{1},
Tsuyoshi Id\'{e}\textsuperscript{2},
Rajiv Joshi\textsuperscript{2},
Steven Holmes\textsuperscript{3}, 
Michael Passow\textsuperscript{2} \\
Tajana Rosing\textsuperscript{1} \\
\textsuperscript{1}University of California San Diego, \textsuperscript{2}IBM Research, \textsuperscript{3}Anderon \\
\textsuperscript{1}\{quzhao,seemami,tajana\}@ucsd.edu \\
\textsuperscript{2}\{tide,rvjoshi\}@us.ibm.com, michael.passow@ibm.com, \textsuperscript{3}Steven.Holmes@anderon.com
}

\renewcommand{\shortauthors}{Zhao et al.}

\title{ Online Open-World Wafer Defect Discovery using Hyperdimensional Computing}

\begin{abstract}
Conventional wafer defect analysis relies on offline, closed-world classifiers that fail to detect novel, unseen defect signatures emerging during continuous semiconductor manufacturing. To address this, we propose \textbf{WaferHDC}, an online open-world wafer defect discovery framework based on hyperdimensional computing (HDC). The memory-centric architecture of WaferHDC enables rapid anomaly detection and grouping of recurring failures into compact hypervector representations. This allows the memory space to dynamically scale with new defect modes without retraining a static classifier or archiving entire raw wafer history. WaferHDC first models normal spatial distributions, then dynamically groups recurring failures into persistent defect memory without labels or predefined class counts. It outperforms the strongest online baseline by 13.3 p.p. in adjusted Rand index and 10.5 p.p. in normalized mutual information on the industry-standard WM-811K dataset. These results demonstrate the effectiveness of WaferHDC for continuous and open-world wafer defect analysis.
\end{abstract}


\maketitle

\section{Introduction}

\begin{figure*}[t]
    \centering
    \includegraphics[width=1\textwidth]{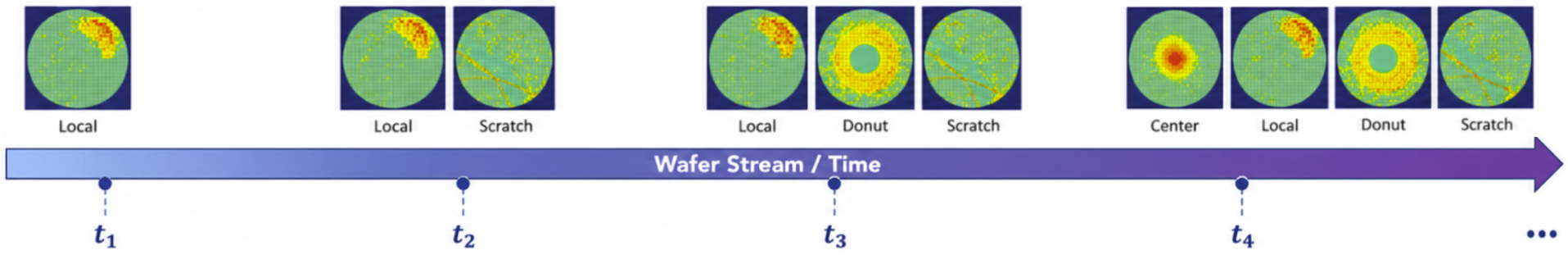}
    \vspace{-8mm}
    \caption{Online open-world wafer defect discovery:
    As wafers arrive, previously unseen defects  may appear, and recurring modes gradually emerge. A monitoring system detects abnormal wafers and organizes recurring patterns without labels. Defect labels are shown only for illustration.}
    \label{fig:stream}
\end{figure*}

Semiconductor wafers undergo a sequence of fabrication and testing steps, and die-level test outcomes are commonly summarized as wafer maps. Image-based patterns in these maps can reveal systematic variation or failures associated with a process step, tool, or handling condition~\cite{bao2024wafer,wu2014wafer,zhu2022optical}. Detecting such patterns is important not only for screening defective wafers, but also for recognizing recurring process problems, supporting root-cause analysis, and preventing yield loss~\cite{ezzat2021graph,zhu2022optical}. Consequently, practical wafer monitoring should go beyond simply flagging defective wafers; it should also identify recurring failure patterns and organize them into meaningful defect modes that can support downstream diagnosis.

\tblue{In real production, wafer monitoring is both online and open-world: wafers arrive sequentially, defect identities and counts are unknown, and new failure patterns may emerge after deployment~\cite{hinder2024one,li2025tensor,frittoli2022deep}. The system must therefore detect deviations from normal behavior and organize recurring anomalies into defect modes as they appear in the stream, as illustrated in Fig.~\ref{fig:stream}.}

Most existing wafer defect analysis methods are developed under offline or closed-world assumptions. Supervised approaches primarily formulate wafer analysis as a classification problem, where a fixed set of defect categories is defined in advance and labeled examples of these categories are available during training~\cite{nag2022wafersegclassnet,bao2024wafer,ko2023novel,mishra2024wafer2spike}. Unsupervised and anomaly-detection methods relax the need for defect labels, but are commonly evaluated on static datasets, either identifying wafers that deviate from normal patterns or clustering a complete collection of wafers after all samples are available~\cite{jo2021one,ezzat2021graph,pleli2024iterative}. Such formulations are valuable for recognizing known failures or detecting abnormal wafers, but they do not explicitly model how previously unseen defect patterns emerge, recur, and accumulate over a continuous production stream. Addressing this issue is challenging for several reasons. First, normal wafer patterns can themselves exhibit substantial variation, making it difficult to characterize normal production behavior. Second, an emerging defect mode may initially be represented by only a small number of wafers and must be distinguished from isolated anomalies before sufficient evidence has accumulated. Third, online discovery requires persistent and efficient memory representation: the system must retain previously observed patterns, accumulate new evidence, and incorporate newly discovered modes online without storing full wafer history.

\tblue{We propose \textbf{WaferHDC}, an online open-world method that unifies wafer defect detection and defect-mode discovery through hyperdimensional computing (HDC)~\cite{kanerva2009hyperdimensional,kleyko2022survey}. WaferHDC models normal production behavior using HDC memory and rejects wafers that are inconsistent with these normal patterns. The rejected wafers are processed by an online discovery mechanism that accumulates similar anomalies, identifies recurring spatial signatures, and promotes them into persistent defect-mode memories. Once a mode is discovered, future wafers can be assigned directly to it, while unfamiliar patterns continue to be accumulated for discovery. WaferHDC requires neither defect labels nor prior knowledge of the number of defect classes during online operation. WaferHDC represents wafers as high-dimensional binary hypervectors that can be compared efficiently using Hamming distance and incrementally combined within an associative memory. While prior HDC systems have mainly used such memories for classification over predefined classes~\cite{verges2025classification,zhao2024multimodalhd,10.1109/IPSN61024.2024.00005}, WaferHDC treats them as an evolving record of production behavior. The same representation can model normal variation, accumulate recurring anomalies, and add new defect modes without storing the full wafer history or retraining a classifier. The main contributions of this work are summarized as follows:}
\begin{itemize}[leftmargin=*]
\item We study wafer defect analysis under an online open-world formulation that jointly considers unknown-defect detection and the continual discovery of recurring defect modes from a wafer stream.
\item We propose \textbf{WaferHDC}, a unified HDC framework that models normal wafer patterns and incrementally discovers and maintains emerging defect modes without defect labels or a predefined number of classes.
\item WaferHDC improves Adjusted Rand Index (ARI) by 13.3 percentage points and Normalized Mutual Information (NMI) by 10.5 percentage points over the strongest existing statistical detectors and streaming-clustering baselines. These results suggest that WaferHDC does not merely fragment anomalous wafers into local microclusters, but instead forms a compact memory of recurring and semantically coherent defect patterns.
\end{itemize}

\section{Related Work}

\begin{table}[t]
\centering
\caption{\tblue{Comparison of representative wafer-map analysis methods.
``Open-world'' handles patterns outside predefined
training classes. ``Online''  causally processes
previously observed wafers. ``Mode discovery'' organizes
unknown wafers into distinct recurring groups.}}
\label{tab:related_comparison}
\scriptsize
\setlength{\tabcolsep}{3pt}
\renewcommand{\arraystretch}{1.12}
\begin{tabularx}{\columnwidth}{@{}Xccc@{}}
\toprule
\textbf{Method} &
\textbf{Open-world} &
\textbf{Online} &
\textbf{Mode discovery} \\
\midrule

WaferSegClassNet~\cite{nag2022wafersegclassnet}
& No & No & No \\

One-Class Wafer Map~\cite{jo2021one}
& Yes & No & No \\

Deep Open-Set Recognition~\cite{frittoli2022deep}
& Yes & No & No \\

Iterative Cluster~\cite{pleli2024iterative}
& Yes & No & Yes \\

Enhanced Open-Set Recognition~\cite{shin2025enhanced}
& Yes & No & No \\

DECOR~\cite{jothiraj2025decor}
& Yes & No & Yes \\

CNN Unknown Defect~\cite{cheon2019convolutional}
& Yes & No & No \\

\textbf{WaferHDC}
& \textbf{Yes} & \textbf{Yes} & \textbf{Yes} \\

\bottomrule
\end{tabularx}
\end{table}

\tblue{Wafer map analysis has been extensively studied for automated defect recognition and process monitoring. Supervised methods formulate the problem as defect classification and have achieved strong performance with statistical methods~\cite{taha2026observational}, deep neural nets~\cite{nag2022wafersegclassnet}, and more recent attention-based architectures~\cite{mohammad2025semiconductor}.} These methods are effective when defect categories are defined in advance and representative labeled examples are available. Such closed-set formulations are fundamentally limited when new process failures produce spatial patterns that were not present during training. 

Anomaly-detection approaches~\cite{frittoli2022deep,shin2025enhanced} and one-class methods~\cite{jo2021one} relax the closed-set assumption by modeling normal wafer behavior and flagging patterns that were not represented during training. \tblue{However, identifying a wafer as abnormal is only the first step toward corrective action. To determine what process issue should be investigated, engineers must recognize whether multiple abnormal wafers exhibit a common spatial signature and therefore may originate from the same recurring failure mechanism. This requires organizing similar anomalies into coherent defect-mode clusters that can be tracked, prioritized, and analyzed. Consequently, an important gap remains between detecting individual unknown defects and discovering the recurring failure modes needed for root-cause diagnosis and process correction.}

\tblue{Unsupervised wafer-analysis methods can organize defects into groups without predefined labels, but most operate offline on a previously collected dataset~\cite{olschewski2021defect,pleli2024iterative,ezzat2021graph,jothiraj2025decor}. As summarized in Table~\ref{tab:related_comparison}, prior methods either detect previously
unseen patterns without organizing them into recurring modes, or discover defect groups only through offline analysis. They therefore do not identify and track an emerging failure signature while production is still ongoing.}

\tblue{WaferHDC addresses this operational gap by processing wafers as they come through the production line and accumulating similar abnormalities as they recur. Repeated wafers with a common spatial signature are consolidated into a persistent defect-mode memory, indicating that the pattern is unlikely to be an isolated event and may reflect an ongoing production issue. This allows recurring failures to be recognized, tracked, and prioritized early enough to support timely root-cause analysis and corrective action.}

\section{WaferHDC Design}
\label{sec:method}

\begin{figure*}[!t]
    \centering
    \includegraphics[width=0.9\textwidth]{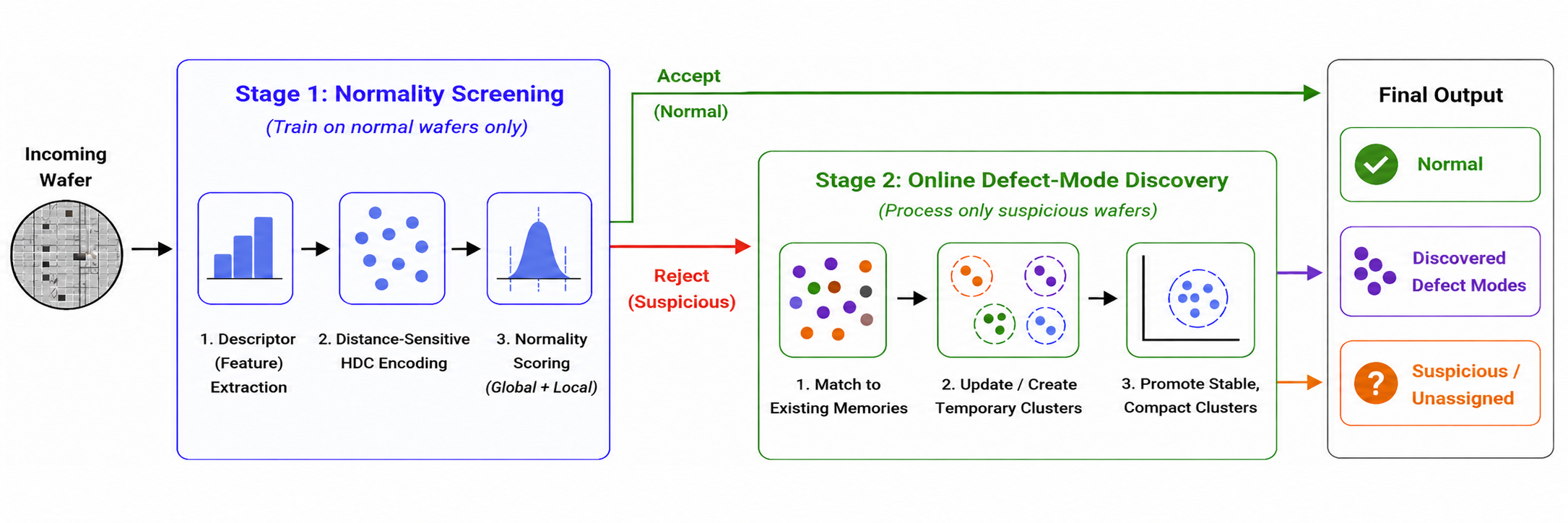}
    \vspace{-10mm}
    \caption{Overview of WaferHDC. Each wafer is encoded as a
    distance-sensitive binary hypervector. Stage~1 determines whether the
    wafer has normal production behavior.
    Suspicious wafers are forwarded to Stage~2, where recurring abnormal
    patterns are accumulated in temporary memories and promoted to
    persistent defect modes after sufficient evidence.}
    \label{fig:overview}
\end{figure*}

\tblue{We consider wafers arriving sequentially as a production stream. WaferHDC processes each incoming wafer in two stages using a shared binary hypervector representation. First, each wafer map is converted into a fixed descriptor of hand-crafted spatial and morphological statistics, including defect density, location, region geometry, concentration, radial distribution, and directional structure; no neural feature extractor is used. The descriptor is then randomly projected, normalized, quantized, and thermometer-encoded into a binary hypervector whose Hamming distance preserves graded differences between wafer patterns. As illustrated in Fig.~\ref{fig:overview}, WaferHDC uses a two-stage design. Stage~1 compares the encoded wafer with multiple normal-pattern memories and a small normal reference bank to determine whether it is consistent with normal production behavior. Wafers identified as abnormal are passed to Stage~2, where they are compared first with persistent defect-mode memories and then with temporary candidate memories. Similar recurring anomalies are accumulated incrementally, while unmatched observations initialize new candidates. A candidate is promoted to a persistent defect mode only after it becomes sufficiently recurrent, internally coherent, and consistently abnormal. In this way, WaferHDC detects unknown defects, discovers recurring failure signatures online, and expands its defect memory without defect labels, full-stream reclustering, or prior knowledge of the number of defect modes.}

\tblue{Before online operation, WaferHDC is provided with a reference set containing only non-defective wafers. No labeled defect examples are available, and the system is not told how many defect modes will appear. This setting differs from closed-set classification, which assumes that all relevant failure categories are represented during training~\cite{frittoli2022deep,shin2025enhanced}, and from offline clustering, which assumes that a complete collection of wafer maps is available for joint analysis~\cite{pleli2024iterative,jothiraj2025decor}. Instead, WaferHDC must recognize normal production behavior, collect evidence for unfamiliar patterns, and identify recurring failure signatures while production is still ongoing.}

\subsection{Distance-sensitive binary wafer encoding}

Given a wafer map $x$, we extract a descriptor
$\mathbf{v}=g(x)\in\mathbb{R}^{d}$ that summarizes its spatial and
morphological characteristics. The descriptor includes complementary
information about defect density, spatial location, connected regions,
local concentration, and directional structure
~\cite{ezzat2021graph,shukla2023embarrassingly,
pleli2024iterative,shin2025enhanced}. In our implementation, $d=202$. The descriptor is projected into a fixed random feature space,
\begin{equation}
    \mathbf{y}=\mathbf{v}\mathbf{R},
    \qquad
    \mathbf{R}\in\mathbb{R}^{d\times P},
\end{equation}
where $\mathbf{R}$ is a Gaussian random matrix. Random projection mixes
the original wafer features into several complementary measurements.
However, directly converting each projected value to binary HDC representation
would retain only angular distance and discard the vector magnitude information, which is essential in distinguishing different defect patterns.

WaferHDC instead quantizes each projected coordinate into a level
$q_p\in\{0,\ldots,B\}$ and represents it using a $B$-bit thermometer
code. A value of \(q_p\) is encoded using \(q_p\) leading ones followed
by zeros. For example, with \(B=5\), levels 1 and 4 are represented as
\(10000\) and \(11110\), respectively. Their Hamming distance is three,
which matches the difference between the two quantized levels:
\begin{equation}
    d_H\!\left(T(q_a),T(q_b)\right)
    =
    |q_a-q_b|.
\end{equation}

The thermometer codes from all projected coordinates are concatenated
to form the wafer hypervector
\begin{equation}
    \mathbf{h}
    =
    T(q_1)\Vert\cdots\Vert T(q_P)
    \in\{0,1\}^{D},
    \qquad D=PB.
\end{equation}
Consequently, Hamming distance between two wafer hypervectors reflects
the accumulated differences between their projected wafer features.
This is important for defect discovery because two patterns may have the
same general direction but differ substantially in defect concentration,
location, or extent. The encoding retains these graded differences while allowing pairwise Hamming distances to be computed using bitwise XOR and population count~\cite{achlioptas2003database,zhang2021fracbnn}.

\tblue{This representation makes both comparison and online learning simple. Similar wafers produce nearby binary hypervectors, so normal prototypes, temporary anomaly candidates, and persistent defect modes can all be compared using the same Hamming-distance operation. Repeated anomalies can also be learned incrementally by updating bit counts and majority prototypes, without storing every previously observed wafer. Different defect orientations are handled during feature extraction rather than through pixel-level alignment. The descriptor includes multi-directional spatial responses, dominant-orientation statistics, and geometry features such as component orientation and major--minor axis structure. Consequently, diagonal, horizontal, and vertical instances of a defect remain characterized by their common morphology, concentration, and directional structure, while meaningful orientation differences are still retained for distinguishing distinct failure patterns.}

\subsection{Stage~1: modeling normal production wafer behavior}

\tblue{Non-defective wafers are not necessarily identical. Benign process
variation, small spatial irregularities, and differences in defect density
can produce several distinct forms of normal behavior. Representing all
normal wafers with a single average pattern may therefore reject valid
process variation.}

Let
$\mathcal{H}_N=\{\mathbf{h}_1,\ldots,\mathbf{h}_{N}\}$
denote the hypervectors encoded from the normal training wafers.
WaferHDC summarizes their diversity using $M$ normal prototypes. The
prototypes are learned through an iterative Hamming-space clustering
procedure. We first initialize $M$ binary prototypes from
$\mathcal{H}_N$. Each normal hypervector is assigned to its nearest
prototype under Hamming distance, forming groups
$\mathcal{A}_1,\ldots,\mathcal{A}_M$. Each prototype is then updated by
bit-wise majority voting over the hypervectors assigned to that group:
\begin{equation}
    p_{m,j}
    =
    \mathbb{I}
    \left[
    \sum_{\mathbf{h}_i\in\mathcal{A}_m}h_{i,j}
    \geq
    \frac{|\mathcal{A}_m|}{2}
    \right].
\end{equation}
Assignment and prototype update are repeated until the assignments
stabilize or a fixed iteration limit is reached. The resulting normal
prototype memory is
\begin{equation}
    \mathcal{M}_N
    =
    \{\mathbf{p}_1,\ldots,\mathbf{p}_M\}.
\end{equation}
Each prototype therefore represents one commonly observed mode of normal
wafer variation rather than averaging all normal wafers into a single
pattern. For an incoming wafer hypervector $\mathbf{h}$, the global normal
distance is its distance to the nearest prototype:
\begin{equation}
    d_g(\mathbf{h})
    =
    \min_{\mathbf{p}\in\mathcal{M}_N}
    d_H(\mathbf{h},\mathbf{p}).
\end{equation}
A small $d_g(\mathbf{h})$ indicates that the wafer resembles one of the
major normal production patterns.

\tblue{$\mathcal{M}_N$ is nevertheless a compressed summary and may only coarsely represent the distribution of non-defect samples. A wafer can therefore be
moderately close to a prototype even when few actual normal wafers
resemble it. To verify local support, WaferHDC also maintains a reference
bank
$\mathcal{R}_N\subseteq\mathcal{H}_N$
containing observed normal hypervectors. Its local normal distance is
defined as:}
\begin{equation}
    d_l(\mathbf{h})
    =
    \frac{1}{k}
    \sum_{\mathbf{r}\in
    \operatorname{kNN}(\mathbf{h},\mathcal{R}_N)}
    d_H(\mathbf{h},\mathbf{r}),
\end{equation}
where
$\operatorname{kNN}(\mathbf{h},\mathcal{R}_N)$
denotes the $k$ closest normal references under Hamming distance.
Intuitively, $d_g$ asks whether the wafer resembles a major normal mode,
whereas $d_l$ asks whether it is supported by actual nearby normal
examples.

Because the nearest-prototype distance and the average-neighbor distance
have different numerical scales, they are normalized.
For \(u\in\{g,l\}\), let \(m_u\) and \(q_u\) denote the median and interquartile range of the corresponding calibration distances. We define:
\begin{equation}
    z_u(\mathbf{h})
    =
    \frac{d_u(\mathbf{h})-m_u}
         {q_u},
    \qquad u\in\{g,l\},
\end{equation}
This robust scaling makes
the global and local signals directly comparable while limiting the
influence of unusually large validation distances. WaferHDC combines the two signals using
\begin{equation}
    s_1(\mathbf{h})
    =
    \max
    \left\{
    z_g(\mathbf{h}),
    z_l(\mathbf{h})
    \right\}.
\end{equation}
Taking the maximum gives a conservative screening rule: a wafer is
considered suspicious when it either differs from all major normal modes
or lacks support from nearby normal observations. The wafer is accepted
as normal when
$s_1(\mathbf{h})\leq\tau_N$;
otherwise, it is forwarded to Stage~2 together with its abnormality
score. The threshold $\tau_N$ is calibrated using only non-defect
validation wafers.

\subsection{Memory for recurring abnormal patterns}

Stage~2 maintains a set of temporary candidate memories
\begin{equation}
    \mathcal{M}_T
    =
    \{C_1^T,\ldots,C_{L_T}^T\},
\end{equation}
where $L_T$ is the current number of candidates and each
$C_i^T$ represents an unresolved abnormal pattern observed in the wafer
stream. A candidate may contain only one unusual wafer when it is first
created. As additional wafers with similar spatial signatures arrive,
they are accumulated into the same memory. The candidate therefore
collects evidence that an abnormal pattern is recurring, without
immediately declaring it to be an established defect mode.

Storing every wafer assigned to a candidate would cause memory usage to
grow continuously with the production stream. WaferHDC instead represents
each candidate $C$ using three incrementally maintained quantities:
\begin{equation}
    C
    =
    \left(
    \mathbf{c}_C,\,
    n_C,\,
    a_C
    \right).
\end{equation}
Here, $n_C$ is the number of wafers accumulated in the candidate,
$\mathbf{c}_C=[c_{C,1},\ldots,c_{C,D}]\in\mathbb{N}^{D}$ is a per-bit
count vector, and $a_C$ is the accumulated Stage~1 abnormality score.
For each bit position $j$,
\begin{equation}
    c_{C,j}
    =
    \sum_{\mathbf{h}_i\in C} h_{i,j},
\end{equation}
so $c_{C,j}$ records how many wafers represented by the memory contain a
one at bit $j$. A newly initialized candidate containing hypervector
$\mathbf{h}$ has
$n_C=1$, $\mathbf{c}_C=\mathbf{h}$, and
$a_C=s_1(\mathbf{h})$. When another wafer hypervector $\mathbf{h}$ is
assigned to the candidate, the memory is updated:
\begin{equation}
    \mathbf{c}_C\leftarrow\mathbf{c}_C+\mathbf{h},
    \qquad
    n_C\leftarrow n_C+1,
    \qquad
    a_C\leftarrow a_C+s_1(\mathbf{h}).
\end{equation}
The update therefore requires only an element-wise addition and does not
require retaining the individual historical wafers.

\tblue{The bit counts provide both a representative pattern and information
about variation within the candidate. When $c_{C,j}$ is close to either
zero or $n_C$, most wafers agree on bit $j$. A count near $n_C/2$
indicates that the assigned wafers disagree at that position. A majority
prototype can therefore be recovered from the counts by setting a bit to
one whenever $c_{C,j}\geq n_C/2$, while the distance from an incoming
hypervector $\mathbf{h}$ to the complete memory is}
\begin{equation}
    d_C(\mathbf{h})
    =
    \frac{1}{n_C}
    \sum_{j=1}^{D}
    \left[
    h_j(n_C-c_{C,j})
    +
    (1-h_j)c_{C,j}
    \right].
\end{equation}
This expression has a direct interpretation:
\begin{equation}
    d_C(\mathbf{h})
    =
    \frac{1}{n_C}
    \sum_{\mathbf{h}_i\in C}
    d_H(\mathbf{h},\mathbf{h}_i).
\end{equation}
Thus, the count-based distance is exactly the mean Hamming distance from
the incoming wafer to all wafers represented by the candidate, even
though those individual hypervectors are no longer stored. WaferHDC also uses the bit counts to measure the internal compactness of
a candidate:
\begin{equation}
    r(C)
    =
    \sum_{j=1}^{D}
    \frac{2c_{C,j}(n_C-c_{C,j})}{n_C^2}.
\end{equation}
A small $r(C)$ indicates that the accumulated wafers agree on most bits
and therefore exhibit a coherent recurring pattern. A large value
indicates that the candidate contains diverse or inconsistent
abnormalities and should not yet be treated as one defect mode. The
average abnormality of the candidate is obtained directly from the
running score as $\bar{s}_1(C)=a_C/n_C$.

The same compact representation is used after a candidate is promoted
into a persistent defect-mode memory. Consequently, WaferHDC can
continuously update both emerging and established defect patterns using
fixed-size memory per mode, rather than storing the complete wafer
history.

\subsection{Recurrence-driven defect discovery}
For each wafer rejected by Stage~1, WaferHDC first compares it with the
persistent defect-mode memories $\mathcal{M}_D$. If the nearest persistent
memory is within the matching threshold $\tau_D$, the wafer is assigned
to that known defect mode and updates its memory. This allows an
established failure pattern to be recognized whenever it appears again.

When the wafer does not match any persistent mode, it is compared with
the temporary candidate memories $\mathcal{M}_T$. If it is sufficiently
close to an existing candidate, it is added to that candidate. Otherwise,
a new temporary memory is initialized. At this stage, the pattern remains
unresolved: one unusual wafer is not enough evidence to declare a new
production failure mode.

After each update, WaferHDC checks whether the candidate has accumulated
enough evidence for promotion. Candidate \(C\) becomes a persistent
defect mode when
\begin{equation}
    n_C\geq n_{\min},
    \qquad
    r(C)\leq\tau_r,
    \qquad
    \bar{s}_1(C)\geq\tau_s.
\end{equation}
The three requirements have direct operational interpretations. The
candidate must have appeared sufficiently often, its wafers must exhibit
a consistent spatial pattern, and the group must remain clearly abnormal
relative to normal production behavior. These checks prevent isolated
outliers, weak Stage~1 rejections, and incoherent mixtures from being
stored as established defect modes.

\tblue{Once promoted, the candidate is moved from $\mathcal{M}_T$ to
$\mathcal{M}_D$. Future matching wafers can then be recognized and added
directly to that persistent memory. Meanwhile, other unfamiliar wafers
continue to form or update temporary candidates. The persistent memory
therefore grows with production, recording recurring failure signatures
without requiring defect labels, a predefined number of classes, or
periodic retraining of a fixed classifier.}

\begin{figure}
    \centering
    \includegraphics[width=1\linewidth]{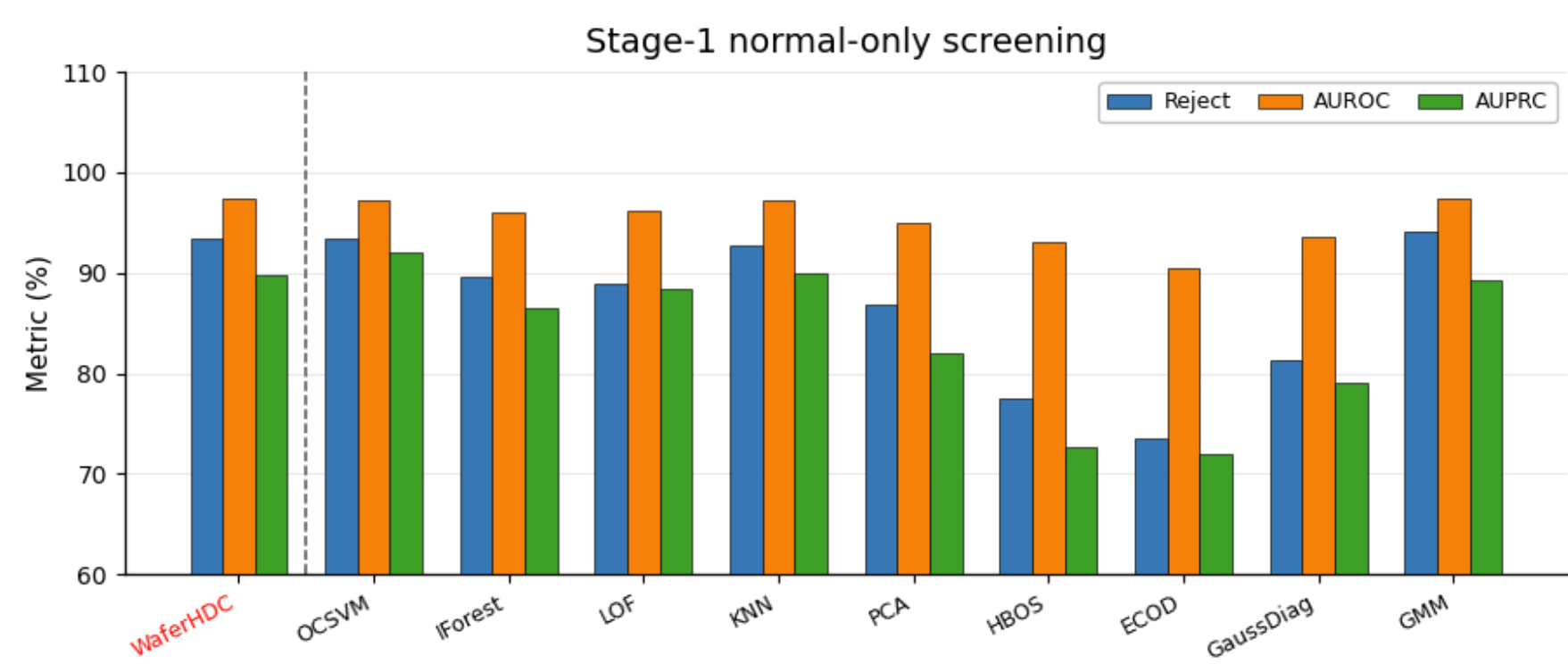}
    \vspace{-5mm}
    \caption{Stage~1 normal-only defect screening on WM-811K. WaferHDC achieves competitive unknown-defect rejection, AUROC, and AUPRC against anomaly-detection baselines.}
    \label{fig:stage1_results}
\end{figure}

\begin{figure}
    \centering
    \includegraphics[width=1\linewidth]{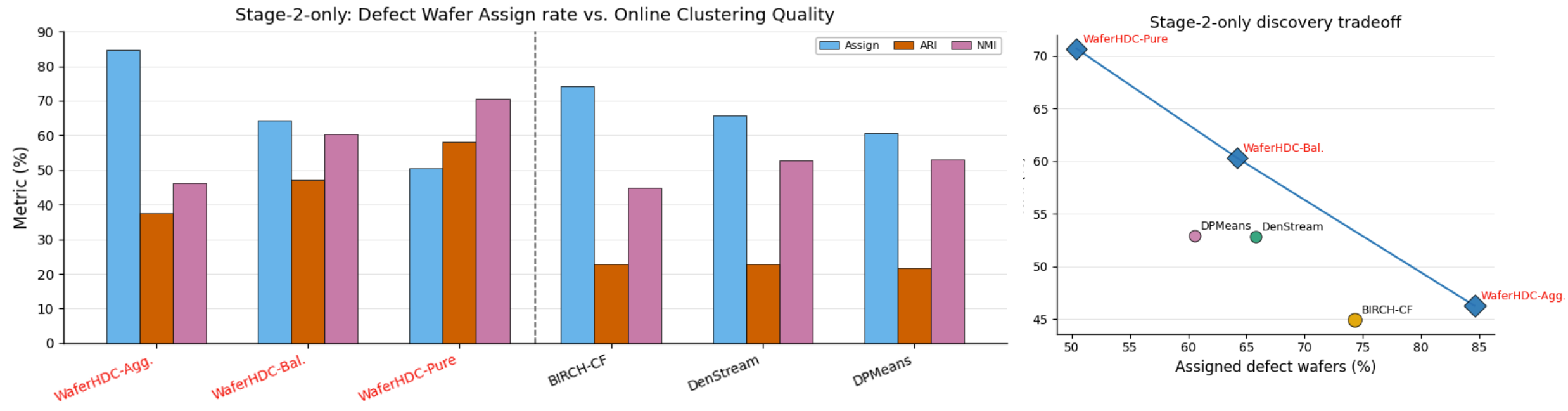}
    \vspace{-5mm}
    \caption{Stage~2-only online defect-mode discovery. Left: assignment coverage and clustering alignment measured by ARI and NMI. Right: the coverage--quality tradeoff in terms of assigned defect wafers and NMI.}
    \label{fig:stage2_results}
    \vspace{-4mm}
\end{figure}

\begin{figure*}[t]
    \centering
    \includegraphics[width=1\linewidth]{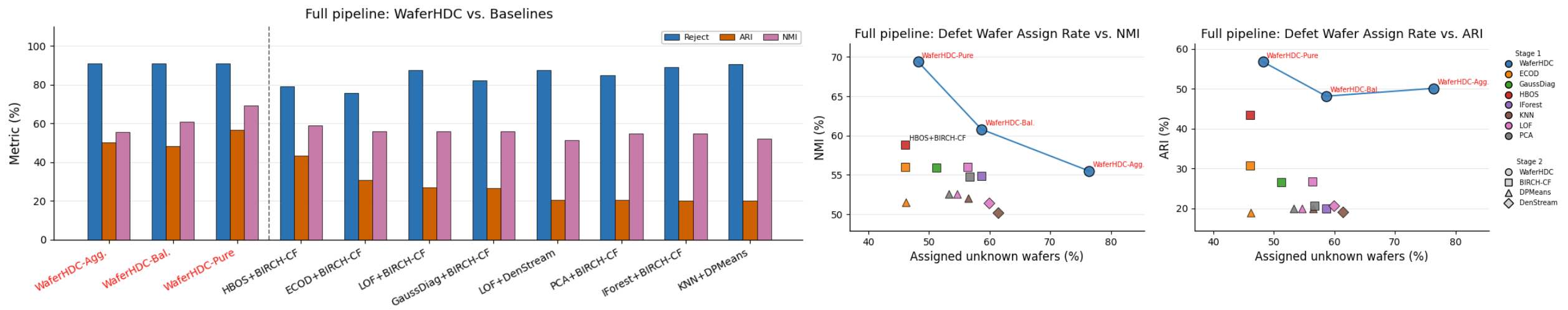}
    \vspace{-7mm}
    \caption{Full-pipeline online open-world defect discovery. Left: defect rejection and clustering alignment for WaferHDC and representative detector--online-clustering baselines. Right: defect assignment rate versus NMI, ARI, with color indicating the Stage~1 detector and marker shape indicating the Stage~2 discovery method.}
    \label{fig:full_pipeline_results}
\end{figure*}

\section{Experimental Evaluation}

\subsection{Experimental Setup}

\noindent\textbf{Baselines:} \tblue{For stage~1, prior wafer studies have explored one-class modeling, nearest-neighbor novelty detection, PCA-based unsupervised analysis, and Gaussian mixture modeling for defect detection and wafer-pattern analysis. We therefore evaluate representative anomaly detectors spanning boundary-based (OCSVM)~\cite{jo2021one}, isolation-based (IForest)~\cite{frittoli2022deep}, local distance and density (KNN and LOF)~\cite{li2022ecod}, reconstruction-based (PCA)~\cite{pleli2024iterative}, and statistical density modeling (HBOS, ECOD, diagonal Gaussian, and GMM)~\cite{han2022adbench}. For Stage~2, we compare with BIRCH-CF~\cite{zhang1996birch}, DenStream~\cite{cao2006density}, and DP-means~\cite{kulis2011revisiting}, covering hierarchical summary-based, density-based stream, and non-parametric clustering, respectively. Recent applications have used these approaches for sequential industrial data, evolving monitoring streams, and dynamic pattern discovery~\cite{maschler2022towards,sedaghatbaf2021dendrift,hanley2024specious}. Unlike recent wafer clustering studies that largely consider offline datasets~\cite{pleli2024iterative}, these methods support incremental cluster formation or a dynamically determined number of clusters. For full-pipeline baselines, each Stage~1 detector is paired with each Stage~2 clusterer, and only rejected wafers are forwarded to online clustering.}

\noindent\textbf{Dataset and protocol.} We evaluate WaferHDC on WM-811K~\cite{wu2014wafer} dataset. The \textit{none} class is treated as non-defect, while the remaining eight classes are unknown defect modes. We use 20\% of the data for testing and 10\% for validation, with at most 5,000 non-defect training wafers. We construct a phased online stream in which normal wafers are distributed throughout the stream and defect modes appear in class-wise phases. Defect labels and the number of modes are unavailable during processing. The Stage~1 threshold is selected from the non-defect validation score distribution, while Stage~2 distance thresholds are calibrated from nearest-neighbor Hamming distances among validation samples rejected by Stage~1; no defect labels are used for threshold selection. All methods receive the same stream.

\noindent\textbf{Metrics:} \tblue{For Stage~1, we report Area Under the Receiver Operating Characteristic curve (AUROC), Area Under the Precision-Recall Curve (AUPRC), and defect wafer rejection rate to evaluate how effectively the system separates abnormal wafers from normal production behavior. For Stage~2 and the full pipeline, we report Adjusted Rand Index (ARI), Normalized Mutual Information (NMI), assigned-defect wafer fraction, and normal contamination. ARI and NMI measure whether the discovered groups align with the underlying recurring defect patterns, indicating how well the method separates distinct failure signatures rather than merely forming arbitrary clusters. The assigned-defect wafer fraction measures the percentage of defect wafers successfully incorporated into a discovered mode and therefore reflects discovery coverage. Normal contamination measures the percentage of non-defect wafers incorrectly absorbed into defect modes, which is important because contaminated modes may mislead engineers during root-cause analysis. Together, these metrics evaluate whether the system discovers coherent and sufficiently complete defect modes while avoiding false process alarms.}

\noindent\textbf{WaferHDC configuration.} We use $P=1024$ and $B=8$, producing 8192-bit hypervectors. Stage~1 uses 10 normal prototypes and $k=10$ local neighbors from at most 2000 normal references. We report \textit{Purity}, \textit{Balanced}, and \textit{Aggressive} variants of WaferHDC, which share the same encoder and Stage~1 model but use different Stage~2 hyperparameters focusing on different aspects.

\subsection{Stage-Wise Evaluation}
\noindent\textbf{Stage~1: Unknown defect detection.} We first evaluate the normal-only screening stage independently. As shown in Fig.~\ref{fig:stage1_results}, WaferHDC achieves strong unknown-defect detection, rejecting 93.48\% of defect wafers with an AUROC of 97.37\%. Its performance is competitive with the strongest classical anomaly detectors across rejection rate, AUROC, and AUPRC, despite using a binary HDC representation. Importantly, Stage~1 is not designed as an isolated anomaly detector. The same distance-sensitive hypervectors and Hamming-space memory representation are directly reused by Stage~2 for online defect-mode discovery. Thus, WaferHDC provides strong normal-only screening while maintaining a unified memory representation for the open-world pipeline.

\textbf{Stage2: Online defect-mode discovery} We next evaluate Stage~2 independently to isolate online defect-mode discovery. As shown in Fig.~\ref{fig:stage2_results}, WaferHDC substantially improves clustering alignment over the online clustering baselines. WaferHDC-Pure achieves 58.07\% ARI and 70.68\% NMI, compared with only about 22--23\% ARI for BIRCH-CF, DenStream, and DP-means. For defect-mode discovery, clustering quality is particularly important: assigning more anomalous wafers is useful only when the resulting groups preserve meaningful defect structure. WaferHDC provides a controllable coverage--quality tradeoff. WaferHDC-Pure favors highly coherent modes, while WaferHDC-Bal. increases assignment to 64.20\% while retaining 60.31\% NMI. Notably, WaferHDC-Agg. assigns a larger fraction of defect wafers than all three baselines while still achieving substantially higher ARI. These results show that recurrence-driven promotion better organizes the abnormal stream into meaningful defect modes, rather than merely maximizing assignment coverage.

\begin{figure}
    \centering
    \includegraphics[width=1\linewidth]{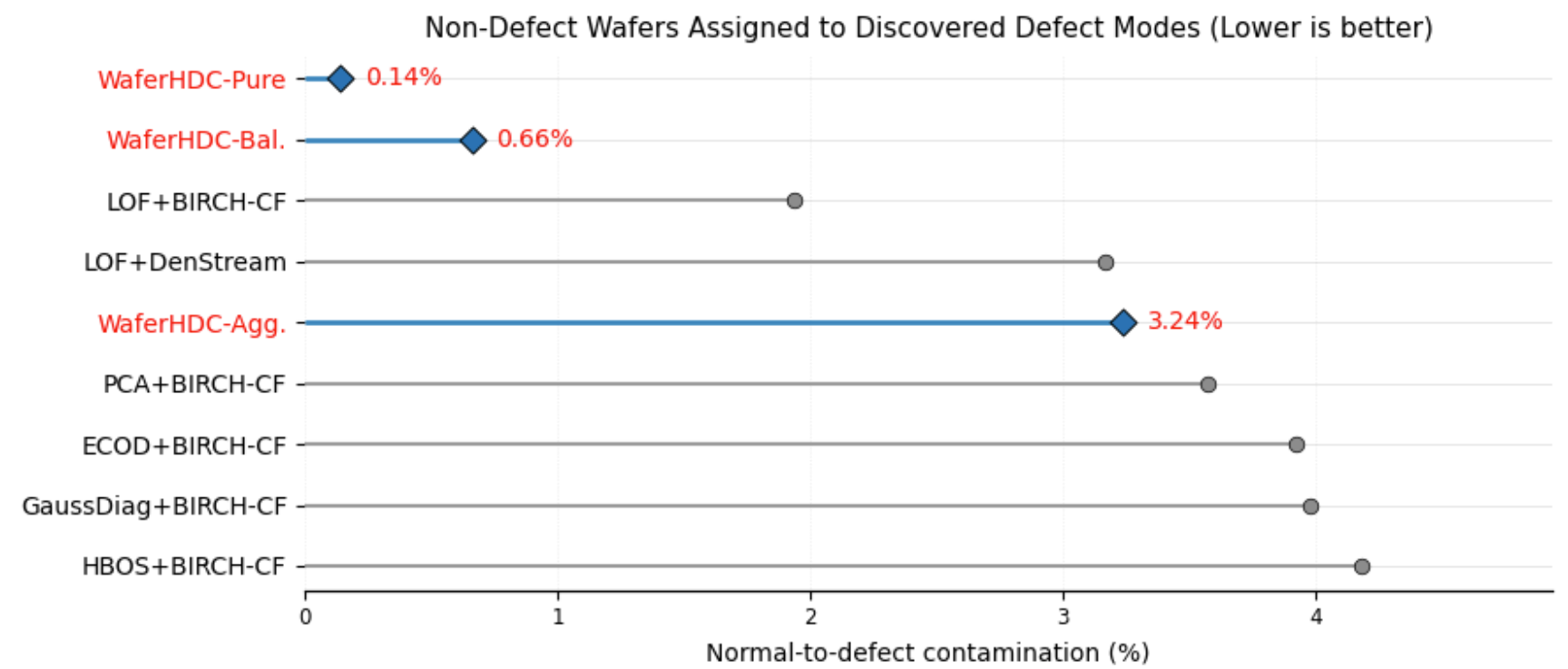}
    \vspace{-7mm}
    \caption{Normal contamination in discovered defect modes}
    \label{fig:contamination}
\end{figure}

\subsection{Full-Pipeline Open-World Discovery}
Finally, we evaluate the complete detector--discovery pipeline under the causal wafer stream. As shown in Fig.~\ref{fig:full_pipeline_results}, WaferHDC maintains its discovery advantage even when Stage~2 receives the imperfect rejected stream produced by Stage~1. WaferHDC-Pure achieves 56.77\% ARI and 69.38\% NMI, improving over the strongest online baseline by 13.3 and 10.5 percentage points, respectively. This shows that the proposed HDC memory framework can recover coherent recurring defect structure despite upstream screening errors. The full-pipeline results also demonstrate a clear coverage--quality tradeoff. WaferHDC-Pure provides the strongest clustering alignment, while WaferHDC-Bal. increases unknown-wafer assignment to 58.66\% and still exceeds the baselines in NMI. WaferHDC-Agg. further increases assignment to 76.42\% and, despite this substantially higher coverage, retains an ARI of 50.11\%, higher than the strongest baseline. The assignment--NMI comparison in Fig.~\ref{fig:full_pipeline_results} further shows that the WaferHDC operating points span a distinct discovery frontier, allowing the system to favor highly coherent modes or broader defect coverage according to monitoring requirements.

We further examine normal contamination in the discovered defect modes. As shown in Fig.~\ref{fig:contamination}, WaferHDC-Pure and WaferHDC-Bal. assign only 0.14\% and 0.66\% of normal wafers to defect clusters, respectively, substantially lower than the baselines. Even WaferHDC-Agg., which emphasizes broader defect coverage, maintains a contamination rate of only 3.24\%. These results indicate that WaferHDC's improved clustering alignment does not come from indiscriminately absorbing rejected wafers; the promoted memories remain well separated from normal production behavior.

\tblue{WaferHDC processes each wafer in approximately \(0.35\)--\(0.81\) ms, corresponding to about \(1{,}200\)--\(2{,}800\) wafers per second. Since industrial wafer probing and testing operate on much longer sub-second to second-scale cycles, WaferHDC is sufficiently fast for online deployment~\cite{advantest_v93000}.}

\section{Conclusion}
Continuous wafer monitoring should move beyond detecting abnormal wafers toward identifying which unknown patterns are becoming recurring failure modes. We introduced \textbf{WaferHDC}, an evolving HDC memory design that unifies normal screening and incremental defect-mode discovery. On WM-811K, WaferHDC achieves competitive defect detection and improves full-pipeline ARI and NMI by 13.3 and 10.5 percentage points over the strongest baseline. These results highlight HDC as an expandable memory framework for continuously learning emerging wafer failures, rather than only recognizing predefined defects.

\begin{acks}
This work was supported in part by PRISM and CoCoSys, centers in JUMP 2.0, an SRC program sponsored by DARPA. This work was also supported in part by the DOE/SC DeCoDe \#84245 led by PNNL, part of MEERCAT and NSF grants \#2003279, \#1911095, \#2112167, \#2052809, \#2112665, \#2120019, \#2211386.
\end{acks}

\bibliographystyle{ACM-Reference-Format}
\bibliography{refs}

\end{document}